\documentclass[conference]{IEEEtran}
\IEEEoverridecommandlockouts%Needed to identify funding in the first footnote.

\usepackage{cite}%lets us cite the bib
\usepackage{amsmath,amssymb,amsfonts}
\usepackage{algorithm}
\usepackage{algpseudocode}
\usepackage{textcomp}
\usepackage{xcolor}
\usepackage{soul}
\usepackage{graphicx} % Required for inserting images
\usepackage{comment}%lets us comment blocks of LaTeX
\usepackage{pdfpages}%lets us add pdfs
\usepackage{xurl}
\usepackage{etoolbox}
\usepackage{hyperref}
\usepackage{csquotes}

\usepackage{tabularx}
\usepackage[most]{tcolorbox}
\usepackage{caption} 
\usepackage{array}
\usepackage{color}
\usepackage{diagbox}
\usepackage{verbatim}
\usepackage{multicol}
\usepackage{multirow}
\usepackage{gensymb}

\newbool{author_sep}
\boolfalse{author_sep}

\makeatletter
\newcommand{\linebreakand}{
  \end{@IEEEauthorhalign}
  \hfill\mbox{}\par
  \mbox{}\hfill\begin{@IEEEauthorhalign}
}
\makeatother

\graphicspath{ {./images/} } %images

\def\BibTeX{{\rm B\kern-.05em{\sc i\kern-.025em b}\kern-.08em
    T\kern-.1667em\lower.7ex\hbox{E}\kern-.125emX}}%not sure

\title{AAD-LLM: Adaptive Anomaly Detection Using Large Language Models\\
\thanks{This material is based upon work supported by the Engineering Research and Development Center - Information Technology Laboratory (ERDC-ITL) under Contract No. W912HZ23C0013. Any opinions, findings and conclusions or recommendations expressed in this material are those of the author(s) and do not necessarily reflect the views of the ERDC-ITL.}
}
\ifbool{author_sep}
{%if author seperated
}
{%if not author seperated
    \author{
        \IEEEauthorblockN{
            Alicia Russell-Gilbert\IEEEauthorrefmark{1}\IEEEauthorrefmark{3},
            Alexander Sommers\IEEEauthorrefmark{1},
            Andrew Thompson\IEEEauthorrefmark{1},\\
            Logan Cummins\IEEEauthorrefmark{1}, 
            Sudip Mittal\IEEEauthorrefmark{1},
            Shahram Rahimi\IEEEauthorrefmark{1},\\
            Maria Seale\IEEEauthorrefmark{2},
            Joseph Jaboure\IEEEauthorrefmark{2},
            Thomas Arnold\IEEEauthorrefmark{2}, Joshua Church\IEEEauthorrefmark{2}
        }
        \IEEEauthorblockA{
            \IEEEauthorrefmark{1}
            \textit{Computer Science \& Engineering} \\
            \textit{Mississippi State University}
            \\\{ar2836, ams1988, agt158, nlc123\}@msstate.edu, \{mittal, rahimi\}@cse.msstate.edu
        }
        \IEEEauthorblockA{
            \IEEEauthorrefmark{2}
            \textit{Engineer Research and Development Center} \\%https://www.erdc.usace.army.mil/
            \emph{Department of Defence}
            \\\{maria.a.seale, joseph.e.jabour, thomas.l.arnold, joshua.q.church\}@erdc.dren.mil
        }
        \IEEEauthorrefmark{3} corresponding author
    }
}

\begin{document}

\maketitle

%abstract--\/

\begin{abstract}
For data-constrained, complex and dynamic industrial environments, there is a critical need for transferable and multimodal methodologies to enhance anomaly detection and therefore, prevent costs associated with system failures. Typically, traditional PdM approaches are not transferable or multimodal. This work examines the use of Large Language Models (LLMs) for anomaly detection in complex and dynamic manufacturing systems. The research aims to improve the transferability of anomaly detection models by leveraging Large Language Models (LLMs) and seeks to validate the enhanced effectiveness of the proposed approach in data-sparse industrial applications. The research also seeks to enable more collaborative decision-making between the model and plant operators by allowing for the enriching of input series data with semantics. Additionally, the research aims to address the issue of concept drift in dynamic industrial settings by integrating an adaptability mechanism. The literature review examines the latest developments in LLM time series tasks alongside associated adaptive anomaly detection methods to establish a robust theoretical framework for the proposed architecture. This paper presents a novel model framework (AAD-LLM) that doesn't require any training or finetuning on the dataset it is applied to and is multimodal. Results suggest that anomaly detection can be converted into a "language" task to deliver effective, context-aware detection in data-constrained industrial applications. This work, therefore, contributes significantly to advancements in anomaly detection methodologies. 
    \\
\end{abstract}

%abstract--/\

\begin{IEEEkeywords}
large language models, LLMs for time series tasks, predictive maintenance, adaptive anomaly detection %keywords
\end{IEEEkeywords}

%intro--\/

\section{Introduction}\label{sect::intro}
We rely on the operation of a variety of engineered systems, which degrade with use and may lead to failures of varying degrees of severity. These failures include unexpected stoppages, product waste, damage to equipment, and bodily harm which have consequences that range from annoying to disastrous. Maintenance practices are crucial to prevent such failures. Condition-based maintenance (CBM) involves performing maintenance actions based on the conditions of a system. Maintenance decisions based on the conditions of a system are often desirable because they allow for proactive and targeted actions. Predictive maintenance (PdM) leverages machine learning to enhance this decision-making process.

PdM becomes more challenging under common real world conditions. Sensor data collected from machines is often non-stationary due to a combination of factors such as varying operational settings and individual machine deterioration \cite{saurav2018online}. This causes heterogeneous relationships between the sensor data and system health; thereby requiring that the normative profile used to identify degradation be updated regularly \cite{Steenwinckel2021FLAGSAM, Steenwinckel2018AdaptiveAD}. We thus do not employ traditional PdM methods, instead employing an adaptive approach. The former cannot account for shifts in sensor data statistical characteristics, while the latter can, all while maintaining high fault detection accuracy.

Unique production system structures and domain-specific constraints necessitate tailored approaches to effectively deploy PdM in diverse industrial settings. The use of expert knowledge into our target use-case facilitate a robust and domain-specific PdM implementation commensurate to these complexities. However, domain-specific knowledge (e.g. process parameter optimal ranges, equipment specifications, etc.) cannot usually be applied across domains and therefore would limit model application across industrial settings. Therefore, retraining or finetuning on the applied dataset with related domain-specific knowledge would typically be required.

It is ideal for models to require infrequent training on a limited dataset yet yield robust generalization capabilities. This is because some critical assets are not allowed to run to failure; thus, event data needed to fine-tune or retrain some machine learning algorithms may be scarce \cite{XIA2018255}. Transferable models that excel in "few-shot" and "zero-shot" scenarios across related domains appear promising. Recent work suggests that pretrained Large Language Models (LLMs) offer noteworthy few/zero-shot capabilities and transferability \cite{LLM_few_shot, LLM_few_shot_health, LLM_few_shot_prog}. Furthermore, the extension of LLMs beyond natural language to the time series domain showcases their broader potential \cite{LLM_in_TS, LLM_zero_shot_TS}. In particular, the application of pretrained LLMs to the problem of anomaly detection for PdM on time series data can improve the transferability of other approaches in data-constrained environments.

To optimize PdM model effectiveness, it is crucial to develop models that are both adaptable and transferable. Adaptable models can adjust to changing conditions, ensuring continued relevance over time. Transferability enables these models to be applied across diverse systems and domains, increasing usability and practicality. By combining adaptability and transferability, PdM models become versatile tools that can evolve with operational environments and be leveraged on a variety of industrial datasets. Furthermore, a model that is optimized to detect anomalies across diverse input sources could enable more synergistic forecasting and foster more collaborative decision-making between the model and plant operators.

The main contributions of this work are as follows:
\begin{itemize}
\item We explore repurposing pretrained LLMs for the PdM use-case. More specifically, pretrained LLMs are explored for use in anomaly detection within manufacturing time series data. Thus, we aim to examine LLMs' efficacy beyond conventional forecasting applications.  
\item We present a novel anomaly detection framework (AAD-LLM) utilizing pretrained LLMs for improved transferability in data-constrained contexts. The improved transferability is shown to reduce the need to retrain between domains and systems. Additionally, the framework is shown to enable the enrichment of input time series data with semantics to deliver more collaborative decision-making between the model and plant operators.
\item We leverage an adaptability mechanism that enables the model to adjust to evolving conditions, consequently enhancing detection accuracy.
\end{itemize}

The remaining sections of this paper are as follows. Section \ref{sect::back} discusses the background and foundational work for our proposed methodology. Section \ref{sect::prior_art} examines the state-of-the-art in LLM time series tasks and adaptive anomaly detection methods. Section \ref{sect::core} provides insight on the AAD-LLM architecture and methodology. Section \ref{sect::results} explains experimental results and implications of findings. Finally, Section \ref{sect::conc} concludes the paper and discusses limitations for future work.

%intro--/\

%background--\/

\section{Background}\label{sect::back}
This section serves as a background for understanding LLMs and adaptive anomaly detection as presented in this paper. It aims to provide key terms, baseline definitions, and relevant mathematical notations that are essential for comprehending the concepts discussed. Additionally, this section briefly discusses the initial stages of our research endeavor. It describes the preliminary investigations conducted to lay the groundwork for our current work.

A \textbf{large language model (LLM)} is trained on sequences of tokens and encodes an auto-regressive distribution, where the probability of each token depends on the preceding ones \cite{LLM_zero_shot_TS}. More simply, an LLM is trained on sequences of words or word pieces, and the output is the likelihood of the next word in a sequence given the previous words (i.e., context-aware embeddings). Each model includes a tokenizer that converts input strings into token sequences. Models like GPT-3 and LLaMA-2 can perform zero-shot generalization, effectively handling tasks without specific training  \cite{LLM_zero_shot_TS}. For this work, we repurpose an LLM for time series anomaly detection while keeping the backbone language model intact \cite{Time_LLM}. A binarization function is then applied to the outputs of the LLM to map them to $\{0,1\}$ to obtain the final predictions. The exact binarization function is use-case specific.

\textbf{Transfer learning} is a machine learning technique where the knowledge gained through one task is applied to a related task  with low/no retraining \cite{10.5555/1803899}. Specifically, in transfer learning, we train a model to perform a specific task on the source domain and then make certain modifications to give us good predictions for a related task on the target domain where data is (usually) scarce or a fast training is needed \cite{Neyshabur2020WhatIB}. For this work, we leverage a pretrained LLMs' text synthesizing and reasoning abilities acquired through training on a source domain by transferring this task knowledge to our PdM use-case. Specifically, we show that pretrained LLMs can effectively predict anomalies in time series data by transferring its text synthesizing and reasoning knowledge to our target manufacturing domain. 

In the context of transfer learning, \textbf{generalizability} refers to the ability of a pretrained model to perform well on new domains as well as new data within the same domain \cite{Stachl2019RepositoryPR}. For example, in manufacturing, the statistical properties of raw material attributes change over time. Therefore, if these variables are used as product quality predictors, the resulting models may decrease in validity. This illustrates that even in the same domain, models may have trouble generalizing from one time point to another \cite{BERGGREN2024112465}. This also illustrates that the issue of generalizability connects to the issue of concept drift \cite{Stachl2019RepositoryPR}. \textbf{Concept drift} is the phenomenon where the statistical properties of a domain changes over time, which can then result in a deterioration of models that have previously been trained within that domain \cite{Stachl2019RepositoryPR, BERGGREN2024112465}. In particular, it can lead to a degradation of performance of static models as they become less effective in detecting anomalies. 

\textbf{Adaptive anomaly detection (AAD)} encompasses techniques that can detect anomalies in data streams or in situations where concept drift is present. These techniques make models capable of automatically adjusting their detection behavior to changing conditions in the deployment environment or system configuration while still accurately recognizing anomalies \cite{Steenwinckel2021FLAGSAM, Steenwinckel2018AdaptiveAD}. For this work, the \textbf{adaptability mechanism} refers to the feature that enables the model's definition of normality and related statistical derivatives to adjust with each new data instance. 

\textbf{Windowing} refers to dividing a time series into smaller, manageable segments, which are then processed individually. Windowing (or \textit{sliding window technique}) is used extensively for anomaly detection in time series data due to its many benefits \cite{w13131862}. For our use-case, dividing the time series into windows helps to preserve local information that might be lost when considering the entire time series as a whole and reduce computational load since models can handle smaller inputs more efficiently.  

A process is said to be \textit{"in statistical control"} if it is not experiencing out of control signals or significant variations beyond normal statistical variations \cite{mcshane2023asq}. \textbf{Statistical process control (SPC)} techniques are commonly used in manufacturing for monitoring sequential processes (e.g., production lines) to make sure that they work stably and satisfactorily \cite{Qiu2018SomePO}. In monitoring the stability of a process, statistical process monitoring (i.e., SPC) plays an essential role \cite{qiu2013introduction, SONG2023109469}. The idea is that processes that are in statistical control are deemed to be \textit{stable} processes \cite{mcshane2023asq}. For this work, stable processes form a baseline for \textbf{normal} process behavior. The selection of SPC techniques are use-case specific. For this work, MAMR is implemented. \textbf{Moving average moving range (MAMR)} charts are plotted for each process variable in the time series dataset as shown in \autoref{fig:fig_meth3}. Upper (UCL) and lower (LCL) control limits for the moving average ($X$) and moving range ($mR$) charts are calculated as follows.

\setlength\parindent{50pt} $X$ Chart:
\begin{equation} \label{eq:xucl}
UCL = \overline{X} + 2.66\overline{R}
\end{equation}

\begin{equation} \label{eq:xlcl}
LCL = \overline{X} - 2.66\overline{R}
\end{equation}

$mR$ Chart:
\begin{equation} \label{eq:rucl}
UCL= 3.27\overline{R}
\end{equation}

\setlength\parindent{14pt} %reset indentation amount
The values $2.66$ and $3.27$ are often used as multipliers for estimating control limits in the MAMR chart. However, these multipliers can significantly widen the control limits, making them less sensitive to minor shifts or variations in the process. Therefore, it is important to analyze historical data to determine the typical variability in the process under consideration and select multipliers that reflect the process's actual behavior while maintaining sensitivity.

Through a thorough review of existing literature, initial data collection, and exploratory data analysis, we aimed to gain insights into the complexities of anomaly detection for PdM and to establish a basis for our subsequent research inquiries. This preliminary work allowed us to uncover initial patterns, trends, and areas of interest, shaping the development of our research framework and guiding the formulation of our research hypothesis for the current work.

\begin{figure}
    \includegraphics[width=0.45\textwidth]{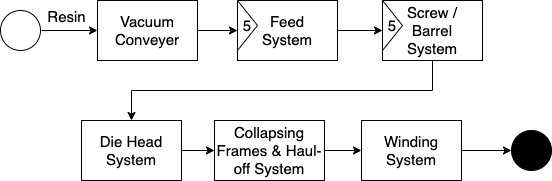}
    \caption{Process flow diagram of major components in our use-case extrusion process. The major components in the extrusion process are in a series configuration. The number of Feed and Screw/Barrel Systems depends on the manufacturing line number and can be 3, 4, or 5.}
    \label{fig:fig_process}
\end{figure}

In a specific use-case study conducted for a leading plastics manufacturing plant, the implementation of anomaly detection algorithms for PdM stood as a cornerstone for the research presented in this paper. An overview of the plastics extrusion process for our use-case can be seen in \autoref{fig:fig_process}. Through anomaly detection, we sought to understand why certain processes failed. Anomalies often indicate underlying factors contributing to process failures. By investigating and correlating anomalies with other process variables, operators and analysts can uncover the root causes of failures. This deeper understanding enables targeted corrective actions to address underlying issues, as well as to better predict future failures.

\begin{figure}
    \includegraphics[width=0.5\textwidth]{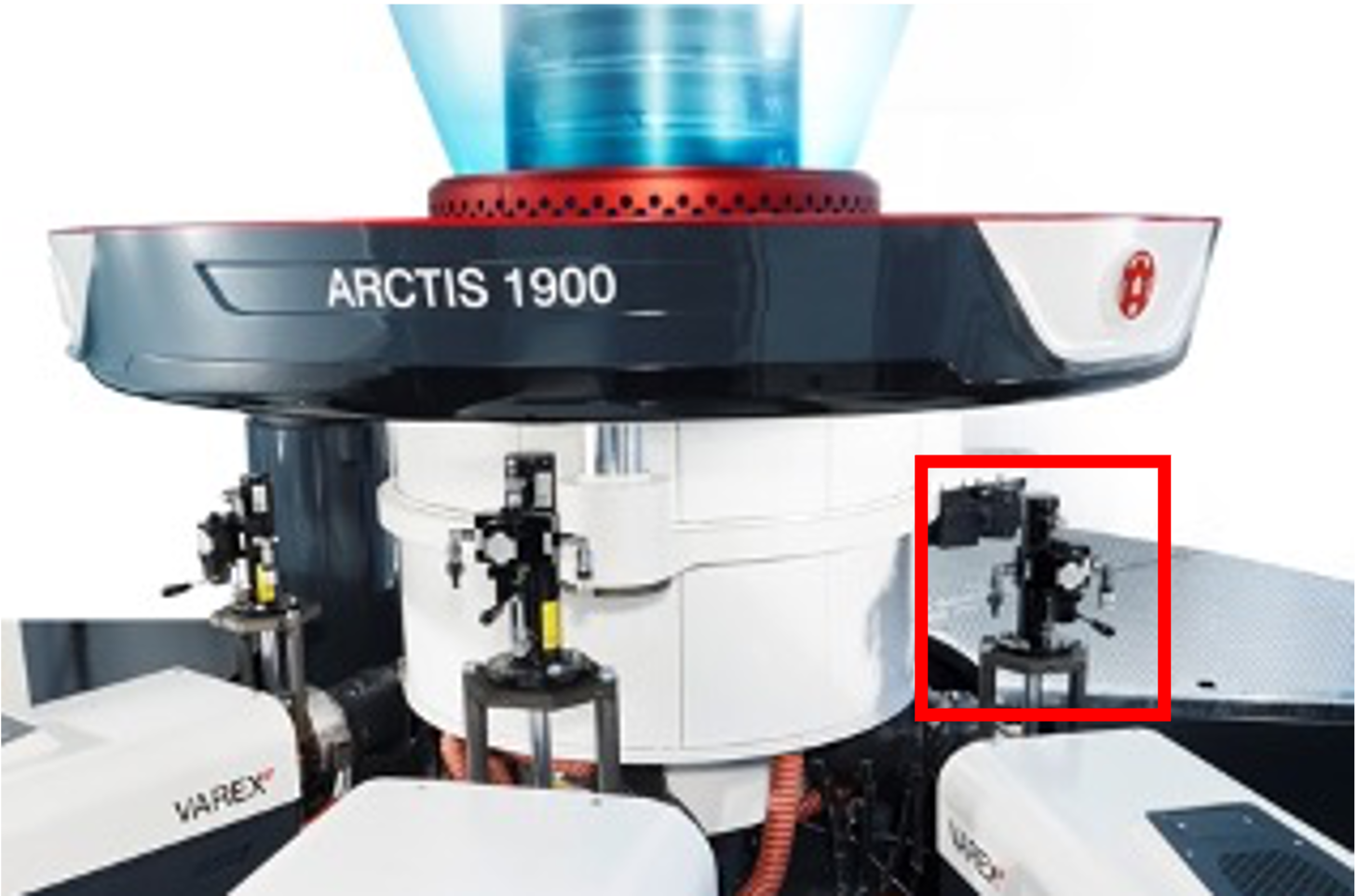}
    \caption{The die head system for our use-case. The screen pack changer is identified by a red box. Within the screen pack changer, screen packs are used to prevent impurities from getting into the extruder together with the resin and thus clogging the die gap. The number of screen packs depend on the number of Screw/Barrel Systems. Each screen pack is arranged between the Screw/Barrel System and the Die Head System. During production, the resin melts flow through the screen pack.}
    \label{fig:fig_screenpack}
\end{figure}

Our investigation focused on screen pack failures since shutdowns due to these failures were well documented in the data. An example of a screen pack changer can be seen in \autoref{fig:fig_screenpack}. For two downtime events with screen pack failure mode, we obtained 65 hours of historical run-to-failure sensor readings (6.5 hours for 5 components for each downtime event). The readings were semi-labeled and for process variables that were deemed good indicators of screen pack failures. These process variables are \textit{Melt Pressure 1}, \textit{Temperature 1}, and \textit{Melt Pressure Differential}. Melt Pressure 1 is the melt viscosity at the screen inlet. Temperature 1 is the melt temperature at the screen pack inlet. Melt Pressure Differential is the melt pressure across the screen pack inlet and outlet. For any of these, sudden spikes from expected profile could signal significant process variable deviations; and therefore, could lead to a screen pack failure.

The data, however, was scarce and noisy; and therefore, presented many challenges in its analysis. While we were able to identify shutdown causes through maintenance logs, we lacked information on where, when, and how failures occurred. To understand why failures happen, it's essential to synthesize this information to uncover root causes, systemic issues, or contributing factors. Additionally, the lack of a general baseline for normal process behavior exacerbated data analysis challenges. The variability in statistical characteristics across processes was influenced by product mix, component age and wear, as well as the heterogeneity of component types.

For the PdM use-case, anomaly detection techniques utilized early on overcame these challenges and showed excellent results. By leveraging domain-specific knowledge, expert rules, sensor readings, and maintenance logs; alongside adaptability methods, we successfully developed a statistics-based predictive model to detect anomalies with $83.3\%$ accuracy. The use case demonstrated the feasibility of anomaly detection for PdM in complex, dynamic, and data-constrained industrial settings. It also highlighted the potential for substantial cost savings, minimized downtime, and enhanced asset reliability. 

The initial approach, while consistent with traditional PdM methods, lacked transferability and multimodality. The statistics-based model was an ensemble approach that relied on domain-specific rules and knowledge to perform well. Domain-specific knowledge (e.g. process parameter optimal ranges, equipment specifications, etc.) cannot usually be applied across domains and therefore would limit model application across diverse industrial settings. Furthermore, the absence of multimodality hindered fusion of diverse input sources (e.g., series data and maintenance logs) for synergistic forecasting. This limitation prevented the enrichment of input sensor readings with semantics, impeding collaborative decision-making between the model and plant operators.

%background--/\

%prior art--\/

\section{Prior Art}\label{sect::prior_art}
%The foundational work for this project is the Time-LLM paper \cite{Time_LLM}. Its code repository can be found \href{https://github.com/KimMeen/Time-LLM/tree/main}{here}.

\subsection{LLMs for Time Series Tasks}
Traditional analytical methods that rely on statistical models and deep learning methods based on recurrent neural networks (RNNs) have dominated the domain of time series forecasting. LLMs, however, have recently emerged in the arena of time series forecasting and have made significant progress in various fields like healthcare, finance, and transportation \cite{LLM_in_TS}. Time-LLM \cite{Time_LLM} proposed a novel framework repurposing LLMs for time series forecasting without requiring any fine-tuning of the backbone model. This was achieved by "reprogramming" time series data inputs for compatibility with LLMs; thereby, converting time series forecasting into a "language" task. An LLM's advanced reasoning and pattern recognition capabilities could then be leveraged to achieve high precision and efficiency in forecasts. Time-LLM was shown to outperform specialized models in few-shot and zero-shot scenarios. 

Similarly, Chronos \cite{Ansari2024ChronosLT} proposed the use of LLMs for time series forecasting. However, it avoided reprogramming the time series data which requires training on each input dataset separately. Instead, time series data was tokenized into a fixed vocabulary via scaling and quantization. The Chronos model outperformed statistical baselines and other pretrained models in both in-domain and zero-shot scenarios across multiple benchmarks. 

LLMTime \cite{LLM_zero_shot_TS} also proposed the use of LLMs for time series forecasting. Rather than requiring learned input transformations or prompt engineering like Time-LLM did, time series data was tokenized like with Chronos but with a different scheme. In fact, for this framework, effective numerical tokenization was essential in ensuring accurate and efficient forecasting by the LLMs. LLMTime outperformed traditional statistical models and models from the Monash forecasting archive. Furthermore, it was competitive with and sometimes outperformed efficient transformer models. 

PromptCast \cite{Xue2022PromptCastAN} also introduced a novel approach to time series forecasting using LLMs. Like Time-LLM, numerical sequences are described and transformed to natural language sentences. However, PrompCast used manually-defined template-based prompting rather than learning input transformations for automatic prompting. While explored for only uni-step forecasting, the results indicated that the PromptCast approach not only achieved performance that was comparable to traditional numerical methods but sometimes even surpassed them. 

These prior works suggest the emergence of multimodal models that excel in both language and time series forecasting tasks. However, these works presented LLMs for use in only time series forecasting and did not explore other time series tasks like anomaly detection. In separate works, however, LLMs have emerged for other time series tasks and have been shown to excel. Time series tasks typically include four principal analytical tasks: forecasting, classification, anomaly detection, and imputation \cite{LLM_in_TS}. 

Zhou et al. \cite{Zhou2023OneFA} introduced a unified framework (referred to as OFA \cite{LLM_in_TS}) using frozen pretrained LLMs for performing various time series analysis tasks. Like Time-LLM, OFA required training the input embedding layer to acquire learned time series representations. However, rather than only time series forecasting, it explored the use of LLMs for univariate anomaly detection. OFA achieved superior or comparable results in classification, forecasting, anomaly detection, and few-shot/zero-shot learning. The TEST method \cite{Sun2023TESTTP} aligns time series embeddings with LLMs to enhance their capability to perform time series tasks without losing language processing abilities. While the exact embedding function was not specified, learning input transformations typically involves neural network training. Therefore, like Time-LLM, TEST also required training the input embedding layer. However, like OFA, TEST explored the use of LLMs for other time series tasks. Compared to state-of-the-art (SOTA) models, TEST demonstrated superior performance on various tasks including univariate time series forecasting, as well as multivariate classification tasks. While achieving good performance on multiple time series tasks, neither OFA or TEST explored multivariate anomaly detection. 

Multivariate analysis allows for joint reasoning across the time series. Joint reasoning enables a model to blend and merge the understanding from different sensors and data sources to make decisions that are impossible when considering data in isolation. For example, in our use-case, the temperature alone may not sufficiently indicate a problem since operators might adjust the temperature to try and maintain material flow despite a screen pack blockage. By monitoring both pressure and temperature, it's possible to detect joint anomaly events that are more indicative of clogging. Furthermore, there were no papers that explored LLMs for the PdM use-case.

\subsection{Adaptive Anomaly Detection}
Advancement in anomaly detection through adaptability has been explored extensively. Traditionally, most AAD algorithms have
been designed for datasets in which all the observations are available at one time (i.e., static datasets). However, over the last two decades, many algorithms have been proposed to detect anomalies in "evolving" data (i.e., data streams) \cite{Salehi2018ASO}. Although the proposed methodology could possibly be modified for data streams, we only focus on static datasets in this paper. 

Machine learning (ML) techniques have been used for AAD implementation and have been shown to improve performance baselines of non-adaptive models in various scenarios such as industrial applications \cite{Singh2023AdaptiveAD}, network security \cite{Xu2023ADTCDAA}, and environmental science \cite{Salehi2018ASO}. However, these techniques only focus on the data themselves. While effective, these approaches may overlook contextual information and domain-specific knowledge crucial for accurate anomaly detection. A system which fuses both ML and semantics improves the accuracy of anomaly detection in the data by reducing the number of false positives \cite{Steenwinckel2018AdaptiveAD, Steenwinckel2021FLAGSAM}. This is because integrating semantics into the anomaly detection process allows for a more comprehensive analysis that considers both the data patterns and their contextual relevance. A system like this would enable more collaborative decision-making between the model and plant operators.

Semantics such as the following could greatly enhance anomaly detection as it provides insight into the severity of the anomaly: \textit{"Domain-specific knowledge indicates that there are correlations between process variables. Specifically, increased melt pressure at the screen pack inlet may lead to increased melt temperature at the screen pack inlet. Additionally, increased melt pressure at the screen pack inlet may lead to decreased melt pressure at the screen pack outlet. If these correlations are observed, it indicates a high level of criticality for potential failures."}. In this case, plant operators may want to imply that if an anomaly is not severe enough, then it is a false positive; and therefore, should not trigger a manual shutdown. Unlike ML models, LLMs can easily integrate this knowledge for the anomaly detection task.

There have been previous works that incorporate expert knowledge with ML algorithms. Steenwinckel et al. \cite{Steenwinckel2018AdaptiveAD} proposed a knowledge graph representation of expert rules which was then transformed directly into matrix form. A reinforcement learning (RL) agent was used as a rule extractor of the matrix (i.e., semantic rule mining) to enable model adaptability.  Adaptability here refers to the mechanism for extracting rules dynamically to better reflect the current state of the environment and ensure continued accuracy. The work, however, was limited as it did not provide a methodology for detecting anomalies in series data using the expert rules as enrichment. 

FLAGS \cite{Steenwinckel2021FLAGSAM} integrated data-driven and knowledge-driven approaches to deliver adaptive, context-aware anomaly detection. The Semantic Mapping module is responsible for enriching the incoming data streams with expert rules and context information. Adaptability here refers to the merging, deleting, or relabeling of anomalies to cope with user-provided feedback; and dynamic rule extracting. FLAGS is an ensemble architecture that uses one ML model to detect anomalies and another that fuses semantics to determine whether they are true anomalies. Although the FLAGS architecture allows for the use of any appropriate ML models, non-LLM models are largely statistical without much innate reasoning \cite{Time_LLM}. 

LLMs, on the other hand, demonstrate advanced abilities in reasoning, data synthesis, and pattern recognition \cite{wang2024enhancing, chu2023leveraging}. Therefore, since pretrained LLMs have been shown to perform well on various time series tasks, leveraging their learned higher level concepts could enable highly precise and synergistic detection across multiple modalities \cite{Time_LLM}.  While traditional statistical models may require more specialized training, LLMs have the ability to perform well with less data and without extensive retraining. This is extremely advantatageous in data-constrained operational settings.

%prior art--/\

%core--\/

\section{Methodology}\label{sect::core}
We assume to have a pretrained LLM as our foundation model. For this work, Meta Llama 3 8B model was chosen since it is an open source instruction-fine-tuned LLM demonstrating SOTA on tasks such as classification, question answering, extraction, reasoning, and summarizing \cite{Zhang2023InstructionTF}. We aim to repurpose this model to detect anomalies in a time series without requiring any fine-tuning. Specifically, this work seeks to test the hypothesis that an LLM can effectively identify anomalies in given time series data for the PdM use-case. 

Additionally, we aim to increase the likelihood of accurately predicting anomalies as the model adapts to temporal changes.  To achieve this, we propose an adaptability mechanism so that with each new data instance, the model is instructed to update its understanding of what constitutes baseline “normal” behavior. By utilizing SPC techniques to establish the baseline, this methodology does not require the model to "learn" normal behavior. Therefore the model maintains high transferability with zero-shot capabilities.

\begin{figure*}
    \includegraphics[width=0.9\textwidth]{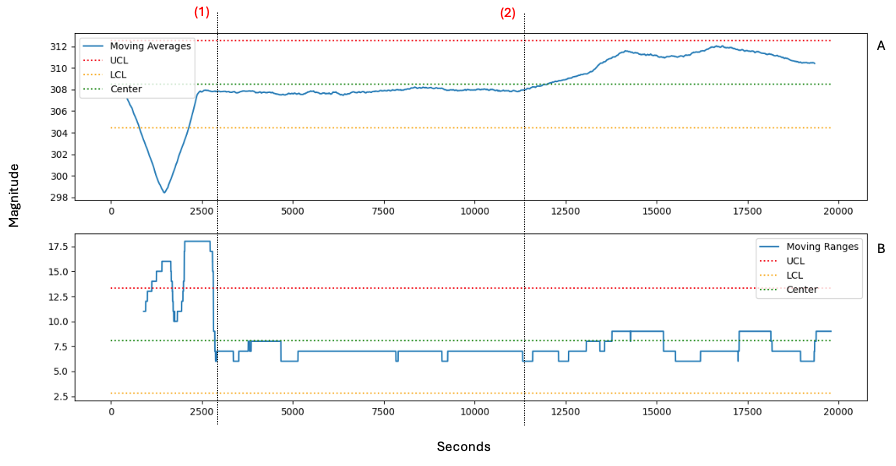}
    \caption{SPC technique of moving average moving range to set control limits for process stability in a query series $Q_i$. Figure A and Figure B are moving average and moving range, respectively. UCL is the defined upper control limit and LCL is the defined lower control limit. Series data points outside of control limits are deemed "out of statistical control" and are labeled as anomalous. Out of control points can be seen before line \textcolor{red}{(1)}. Points between lines \textcolor{red}{(1)} and \textcolor{red}{(2)} represent a stable process.  Points after line \textcolor{red}{(2)} also represent a stable process, however, they are trending towards out of control. These points, therefore, are potentially problematic. AAD-LLM is applied to all points within control limits to enhance anomaly detection.}
    \label{fig:fig_meth3}
\end{figure*}

We then apply a binarization function to the LLM's output to accurately classify the time series as anomalous/non-anomalous. The exact binarization function is use-case specific as described in Section \ref{sect::bin}. The backbone of the pretrained LLM is frozen. The architecture of our proposed framework, referred to as AAD-LLM, incorporates the following main elements: (1) domain-specific knowledge and rules, (2) a comparison dataset $C$, (3) a pretrained and frozen LLM, and (4) an output mapping function. In the following paragraphs, we explain the model architecture in greater detail.

\subsection{Data and Analysis}
For two consecutive downtime events in November of 2023, we obtained 65 hours of run-to-failure sensor readings as described in  Section \ref{sect::back}. Python scripts were used to preprocess this data using the SPC techniques presented previously. Additionally, Python scripts were used to develop the model architecture (AAD-LLM) and to evaluate its performance. Furthermore, a publicly available dataset \cite{skab} was chosen to be implemented by the proposed methodology to compare with selected benchmarks. 

The Skoltech Anomaly Benchmark (SKAB) is a dataset designed for evaluating the performance of anomaly detection algorithms. The benchmark includes 34 labeled datasets of signals captured by several sensors installed on the SKAB testbed. The SKAB testbed was specifically developed to study anomalies in a testbed. The focus of this work is to develop methods for detecting anomalies in these signals, which can be relevant for various applications. 

We implemented AAD-LLM on the SKAB dataset for both valves. The valves are \textit{valve1} which is the outlet of the flow from the pump; and \textit{valve2} which is the flow inlet to the pump. A description of the columns in each dataset is as follows \cite{skab}.
\begin{itemize}
    \item datetime - Dates and times when the value collected
    \item Accelerometer1RMS - Vibration acceleration (g units)
    \item Accelerometer2RMS - Vibration acceleration (g units)
    \item Current - The amperage on the electric motor (Ampere)
    \item Pressure - The pressure in the loop after the water pump (Bar)
    \item Temperature - The temperature of the engine body ($\degree C$)
    \item Thermocouple - The temperature of the fluid in the circulation loop ($\degree C$)
    \item Voltage - The voltage on the electric motor (Volt)
    \item RateRMS - The circulation flow rate of the fluid inside the loop (Liter per minute)
    \item anomaly - If the point is anomalous (0 or 1)
    \item changepoint - If the point is a changepoint (0 or 1)
\end{itemize}

\begin{figure*}
    \includegraphics[width=0.95\textwidth, height=3.0in]{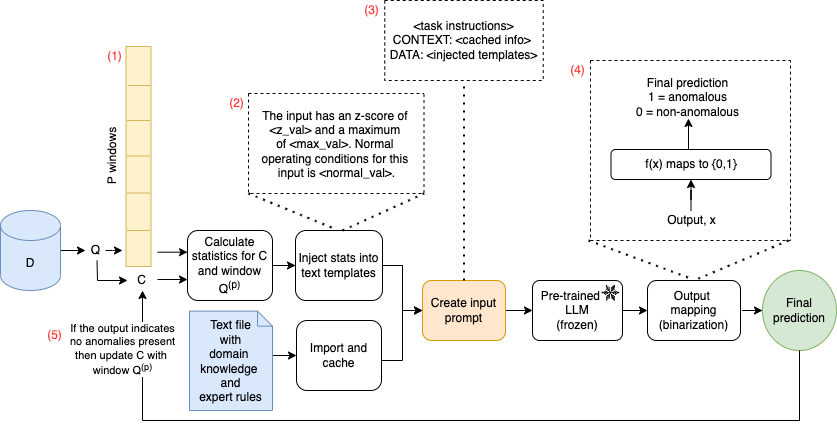}
    \caption{The model framework of AAD-LLM. Given an input time series $Q$ from the dataset $D$ under consideration, we first preprocess it using SPC techniques. Then \textcolor{red}{(1)} $Q$ is partitioned into a comparison dataset $C$ and query windows $Q^{(p)}$, where $p \in P$ and $P$ is the number of segmented windows. Next, statistical derivatives for $C$ and $Q^{(p)}$ are calculated and \textcolor{red}{(2)} injected into text templates. These templates are combined with task instructions to create the input prompt. To enhance the LLM's reasoning ability, \textcolor{red}{(3)} domain context is added to the prompt before being fed forward through the frozen LLM. The output from the LLM is \textcolor{red}{(4)} mapped to $\{0,1\}$ via a binarization function to obtain the final prediction. \textcolor{red}{(5)} Updates to $C$ are determined before moving to the next $Q^{(p)}$.}
    \label{fig:fig_meth1}
\end{figure*}

The anomaly column contains the labels. A Mann–Whitney–Wilcoxon test (or Mann-Whitney U test) was used to determine whether any of the features in the dataset affected the labels. It was determined with $95\%$ confidence that only \textit{Accelerometer1RMS}, \textit{Accelerometer2RMS}, \textit{Temperature}, \textit{Thermocouple}, and \textit{RateRMS} affected the labels. Therefore, only these were input into AAD-LLM to make the predictions. Doing this decreased computational time and improved accuracy, precision, recall, and F1 score.

\subsection{Domain Context and Text Templates}\label{sect::proto} 
To facilitate collaboration with plant operators, we first construct a domain-specific context file, which enables the LLM to understand the context of our time series data. This file incorporates expert rules, domain-specific knowledge, and constraints to define acceptable ranges of process variable variations, guide feature selection, and highlight causal relationships among variables. Real manufacturing data often contains many sensor readings. For our use-case, there are 580 sensors per line but operators can correlate them with failure modes. Furthermore, a change in raw materials often requires a significant change in use-case process parameter values but staff polymer scientists know related thresholds. By leveraging this information, algorithms can refine thresholds, select relevant features, and identify interactions, ultimately enhancing anomaly detection performance. This domain context is gathered in a context file, then imported and cached for improved performance and resource efficiency.

%trends%
To enable structured understanding and improved performance, we create a set of text templates with placeholders for statistical values. Once actual data becomes available, these placeholders will be replaced through data injection. The templates are designed to align with specific statistical derivatives (mean, standard deviation, and maximum as detailed in Section \ref{sect::back}) important for anomaly detection. By combining mean and standard deviation as a z-score, LLMs can focus on reasoning without performing calculations. Statistical derivatives for both normal system behavior and the query window under consideration are injected into their respective text templates as further explained in Section \ref{sect::prompt}. The injected text templates guide the LLM's reasoning, enhancing its performance in detecting anomalies.

\subsection{Initializing the Comparison Dataset} 
Defining normal process behavior is essential for effective anomaly detection. Establishing normal process behavior provides a baseline against which anomalies can be compared and detected. Without a clear understanding of normal conditions, it's challenging to detect abnormal patterns. A well-defined normal process behavior also enables algorithms to differentiate between regular variations and true anomalies, reducing false positives and improving issue identification. However, establishing a universal definition of normalcy is difficult due to the complexity and variability inherent in different manufacturing processes.

Considering the aforementioned, the next step of our proposed algorithm is to initialize datasets $C_i$ that represent normal process behavior. From the dataset D under consideration, a multivariate time series instance $Q\in \mathbb{R}^{N\times T}$ is partitioned into $N$ univariate time series where $N$ is the number of input variables and $T$ is the number of time steps. This is done so that each input variable is processed independently \cite{Time_LLM}. Each $i^{th}$ series $Q_i, i \in N$, is preprocessed using SPC techniques as shown in \autoref{fig:fig_meth3}. Time series points deemed "out of statistical control" are labeled as anomalous and filtered out of $Q_i$ before further processing. SPC is applied again after the first set of outliers (or anomalies) are removed. This is done to ensure extreme values do not affect control limits. Therefore, it can be assumed that time series $Q_i$ represents a stable process. We use this assumption in initializing our comparison dataset $C_i$ as our baseline for normal behavior as explained in the next section. The idea is that once the comparison dataset is initialized, the model then updates its understanding of normalcy as each new query window is ingested.

\subsection{Windowing and Prompting}\label{sect::prompt}
Rather than processing the entire time series at once, $Q_i$ then undergoes windowing as shown in \autoref{fig:fig_meth1}. For each $i \in N$, windowing divides time series $Q_i$ into $P$ consecutive non-overlapping segments of length $L$, $Q_i^{(P)} \in \mathbb{R}^{P \times L}$. By analyzing data within sliding windows, anomaly detection can focus on smaller segments of the time series data. This provides a more granular and detailed view of abnormal patterns. Processing the entire time series as a single entity might obscure localized anomalies within the data. Finally, for each $i \in N$, a baseline dataset $C_i \in \mathbb{R}^{1 \times L}$ of normal behavior is defined as the first $Q_i$ window.  

\setlength\parindent{12pt}
The selection of optimal statistical derivatives is crucial for effective anomaly detection algorithms. By choosing the right derivatives, algorithms can focus on relevant data features, enhancing accuracy and reducing false positives. Different industries and applications have distinct anomaly detection requirements, necessitating customization through appropriate statistical derivatives. In this study, as stated previously in Section \ref{sect::proto}, z-score and maximum are the selected statistical derivatives.

Calculating the statistical derivatives is the next step in our proposed algorithm. For each $i \in N$, statistical derivatives for both the current $Q_i$ window $Q_i^{(p)}$ where $p \in P$, and $C_i$ are calculated and then injected into the text templates. Prompts are then created via prompt engineering and combined with the templates for $Q_i^{(p)}$ and $C_i$ for each $i \in N$. To further enrich the inputs, the domain context is added to the prompt before being fed forward through the frozen LLM. A prompt example is shown in \autoref{fig:prompt_text}. For our methodology, the domain context was manually restructured from the "raw" domain context to better guide the LLM's decision making, thereby enabling more consistent predictions. Effective prompt engineering and domain context structure is essential in ensuring accurate, context-aware anomaly detection.

\setlength\parindent{12pt} %otherwise indention goes away
\subsection{Output Binarization Function and Updating $C_i$}\label{sect::bin}
Upon creating the prompt, it is fed forward through the frozen LLM to obtain the outputs. Lastly, we apply a binarization function to the outputs to map them to $\{0,1\}$ to get the final classification ($0=$ non-anomalous, $1=$ anomalous). The exact binarization function is use-case specific. For our use-case, one anomaly alone does not sufficiently indicate a problem. To avoid false positives that trigger an unnecessary shutdown, our binarization function only maps to $1$ if anomalies in the output are correlated as indicated by domain-specific knowledge. Let $x$ be the LLM output. Then

\begin{equation}
%\[
     f(x) =
    \begin{cases} 
        1, & \text{if anomalies in } x \text{ are correlated} \\
        0, & \text{otherwise}
     \end{cases}
%\]
\end{equation}

The final classification is what is used for determining updates to $C_i$ before moving to the next $Q_i^{\left(p\right)}$. If the output prediction indicates no anomalies in $Q_i^{\left(p\right)}$, window $Q_i^{\left(p\right)}$ series data is combined with the preceding windows series data to gradually refine the dataset of what constitutes "normal" behavior $C_i$. Therefore, for each $i \in N$, $C_i$ is guaranteed to be representative of normal behavior and is constantly evolving.

%begin textbox
\begin{center}
    \begin{minipage}{9cm}            
        \begin{tcolorbox}[enhanced,attach boxed title to top center={yshift=-1mm,yshifttext=-1mm},
                colback=green!10!white,colframe=gray!90!black,colbacktitle=gray!80!black, left=0.1mm, right=0.5mm, boxrule=0.50pt]
        %TEXT                
        Given the following context and data, determine whether there are any anomalies present.\\
        \textbf{CONTEXT:} \textcolor{red}{$<$cached info$>$} \\
        \textbf{DATA:} The following sensor data was collected over the last 15 minutes and represent current process conditions. Temperature 1 has a maximum of \textcolor{red}{$<$val$>$} and a z-score of \textcolor{red}{$<$val$>$}. Normal operating conditions for Temperature 1 is \textcolor{red}{$<$val$>$}. Melt Pressure 1 has a maximum of \textcolor{red}{$<$val$>$} and a z-score of \textcolor{red}{$<$val$>$}. Normal operating conditions for Melt Pressure 1 is \textcolor{red}{$<$val$>$}. Melt Pressure Differential has a maximum of \textcolor{red}{$<$val$>$} and a z-score of \textcolor{red}{$<$val$>$}. Normal operating conditions for Melt Pressure Differential is \textcolor{red}{$<$val$>$}.

        \end{tcolorbox}
    \end{minipage}
\end{center}
\vspace{0mm}
\captionof{figure}{Prompt example. \textcolor{red}{$<$cached info$>$} is the domain context information. \textcolor{red}{$<$val$>$} are calculated statistical derivatives injected into respective text templates. Note that although each $Q_i$ is processed independently, prompts include text templates for all $i \in N$ where $N$ is the number of input variables in instance $Q$ from the dataset $D$ under consideration. Therefore, multivariate anomaly detection is explored. }\label{fig:prompt_text}
\vspace{4mm}
%end textbox

\setlength\parindent{12pt}
\subsection{Adaptability Mechanism}
In addition to $C_i$ constantly updating as each new query window is ingested, the process of re-initializing $C_i$ is done for each new instance $Q$. This continuous redefining of the normal baseline enables the model to progressively refine its knowledge in response to shifts in the system’s operational conditions process after process. Therefore, the model is enabled to maintain an up-to-date and broad perspective of normality. 

%should go in results
\begin{table}[!h]
    \begin{center}
    \begin{tabular}{| c | c | c |}
    \hline
    \textbf{Metric} & \textbf{Use-Case dataset} & \textbf{SKAB dataset} \\
    \hline
    \hline
    Accuracy & 0.706612 & 0.584337 \\
    \hline
    Precision & 0.881481 & 0.467391 \\
    \hline
    Recall & 0.683908 & 0.682540 \\
    \hline
    F1 score & 0.770227 & 0.555839 \\
    \hline
    \end{tabular}
    \caption{AAD-LLM evaluation metrics.} \label{table:tab_results1}
    \end{center}
\end{table}

%core--/\

%results--\/

\section{Results And Discussion}\label{sect::results}
\begin{table*}[!t]
    \begin{center}
    \begin{tabular}{| c | c | c | c | c | c | c | c |}
    \hline
    \textbf{Algorithm} & \textbf{F1} & \textbf{FAR, \%} & \textbf{MAR, \%} & \textbf{No Training or Finetuning} & \textbf{Multimodal} \\
    \hline
    \hline
    Perfect detector & 1 & 0 & 0 & & \\
    \hline
    LSTMCaps \cite{Elhalwagy_2022} & 0.74 & 21.5 & 18.74 & no & no \\
    \hline
    MSET \cite{mset} & 0.73 & 20.82 & 20.08 & no & no \\
    \hline
    LSTMCapsV2 \cite{Elhalwagy_2022} & 0.71 & 14.51 & 30.59 & no & no \\
    \hline
    MSCRED \cite{Zhang_Song_Chen_Feng_Lumezanu_Cheng_Ni_Zong_Chen_Chawla_2019} & 0.7 & 16.2 & 30.87 & no & no \\
    \hline
    Vanilla LSTM \cite{filonov2016multivariateindustrialtimeseries} & 0.67 & 15.42 & 36.02 & no & no \\
    \hline
    Conv-AE \cite{conv_ae}	& 0.66 & 5.58 & 46.05 & no & no \\
    \hline
    LSTM-AE \cite{lstm_ae} & 0.65 & 14.59 & 39.42 & no & no \\
    \hline
    \textbf{AAD-LLM} & \textbf{0.56} & \textbf{47.6} & \textbf{31.7} & \textbf{yes} & \textbf{yes} \\
    \hline
    LSTM-VAE \cite{Bowman2015GeneratingSF} & 0.56 & 9.2 & 54.81 & no & no \\
    \hline
    Vanilla AE \cite{Chen2017OutlierDW} & 0.45 & 7.55 & 66.57 & no & no \\
    \hline
    Isolation forest \cite{4781136} & 0.4 & 6.86 & 72.09 & no & no \\
    \hline
    Null detector & 0 & 100 & 100 & & \\
    \hline
    \end{tabular}
    \caption{Best outlier detection scores for each anomaly detection method implemented on the SKAB dataset, sorted by F1 score \cite{Elhalwagy_2022}. A selection of NNs and ML based fault detection methods were chosen to compare on the benchmarks. Multimodality allows for the enriching of input series data with semantics to enable more collaborative decision-making between the model and plant operators. For this work, multimodality refers to a model being optimized to detect anomalies across both time series data and text. A model that requires no training or finetuning on the data it is applied to is conidered transferable with zero-shot capabilities. Unlike all other methods, AAD-LLM is not trained or finetuned on the dataset it is applied to and is multimodal without requiring any additional strategies.} 
    \label{table:tab_results2}
    \end{center}
\end{table*}

Our brief results are shown in Table \ref{table:tab_results1}. For the use-case dataset, the model achieved an accuracy of $70.7\%$, indicating the proportion of correctly classified instances among all instances was good. Furthermore, an F1 score of $77.0\%$ signifies that the model achieves a balanced performance in terms of precision and recall. This suggests that the model effectively balances correctly identifying both anomalous and non-anomalous series data, offering a robust performance overall. For the SKAB dataset, the model achieved an accuracy of $58.4\%$. While an accuracy of $58.4\%$ is better than random guessing (50\% accuracy), it indicates that the model's prediction performance may be limited. However, upon inspecting the LLM's output for the last input series, z-score comparison errors caused the accuracy to decrease from $84.2\%$ to $42.1\%$. Comparison errors affected the benchmark dataset performance more than the use-case dataset performance. More information regarding these comparison errors can be found in Section \ref{sect::conc}

Table \ref{table:tab_results2} summarizes the scores for algorithms on 3 application benchmarks using the data files in SKAB, sorted by F1 score. For F1 score, bigger is better. For both FAR and MAR, less is better. AAD-LLM ranks $8^{th}$ among 11 algorithms in F1 score, $5^{th}$ in MAR, and last in FAR. Investigation of the LLM's output for the last input series suggests that the FAR score was negatively affected by z-score comparison errors.

This study provides compelling insights into the effectiveness of leveraging LLMs for anomaly detection tasks, particularly in PdM. Results show the successful repurposing of LLMs for anomaly detection, demonstrating their ability to detect anomalies in time series data accurately, even in complex operational settings. These findings confirm the feasibility and strong performance of LLMs in anomaly detection tasks, showcasing their versatility and robustness. The research supports the use of LLMs for anomaly detection in PdM, underlining their capability and potential in handling challenges in time series anomaly detection, especially in data-constrained industrial applications. This study significantly advances anomaly detection methodologies.

%results--/\

%discussion--\/

%\section{Discussion}\label{sect::disc}
%\input{component_subject_sections/Discussion/discussion}

%discussion--/\

%limitations--\/

%\section{Limitations}\label{sect::limits}
%\input{component_subject_sections/Limitation/limitation}

%limitations--/\

%recomendations and future work--\/

%\section{Recommendations and future work}\label{sect::rec_fw}
%\input{component_subject_sections/Recommendations-Future_Work/rec_fw}

%recomendations and future work--/\

%conclusion--\/

\section{Conclusion And Future Work}\label{sect::conc}
Assuming that domain-specific knowlege is available for leveraging, AAD-LLM shows promise in repurposing LLMs for anomaly detection. This is because it allows for greater transferability of anomaly detection models and for enriching time series data with domain-specific knowledge to enable more collaborative decision-making between the model and plant operators. Furthermore, the adaptive mechanism enables the model to progressively refine its understanding of baseline behavior in response to shifts in the system's operational conditions over time. Therefore, the model can maintain an up-to-date perspective of normal system behavior in order to make accurate predictions despite the dynamic nature of operational settings. By utilizing SPC techniques to establish the baseline, this methodology does not require
the model to ”learn” normal behavior. Therefore the model maintains high transferability with zero-shot capabilities. The results suggest that anomaly detection can be converted into a language task to deliver effective, context-aware detection in data-sparse industrial applications. 

Although evaluation metrics demonstrate proof of concept, performance can be improved. The LLM we used was not consistent in accurately making comparisons between statistical derivatives and did not perform calculations well. This lead to a decrease in model performance. Further work should explore adding a Retrieval-Augmented Generation (RAG) pipeline to retrieve relevant mathematical information or formulas from external knowledge sources; thus aiding the model in performing comparisons and math calculations more accurately. Additionally, the framework requires manually restructuring the "raw" domain context to better guide the LLM's decision making; thereby enabling it to make more consistent predictions. This restructured domain context does not read in the way real maintenance logs or operator instructions would naturally read. Further work should explore training a neural network for automatic domain context sentence restructuring. Finally, AAD-LLM was applied to only static datasets to better understand how processes failed after the failure had already occurred. However, further work should explore extending the methodology to data streams for online anomaly detection.

%conclusion--/\

%notes--\/%remove this when finishing the paper

%\section{Notes}\label{Notes to reader}
%\input{component_subject_sections/Notes/Notes}

%notes--/\

%bib settings--\/

\bibliographystyle{ieeetr} % We choose the "plain" reference style
\bibliography{ref} % Entries are in the refs.bib file

%bib settings--/\

\end{document}